\documentclass[conference]{IEEEtran}
\IEEEoverridecommandlockouts
\usepackage[left=1.62cm,right=1.62cm,top=1.9cm]{geometry}
\usepackage{cite}
\usepackage{amsmath,amssymb,amsfonts}
\usepackage{algorithm}
\usepackage{algorithmicx}
\usepackage{algpseudocode}
\usepackage{graphicx}
\usepackage{textcomp}
\usepackage{xcolor}
\usepackage{float}
\usepackage{subcaption}
\usepackage{changepage}
\usepackage{titlesec}

\usepackage{caption} % add the caption package
\def\BibTeX{{\rm B\kern-.05em{\sc i\kern-.025em b}\kern-.08em
    T\kern-.1667em\lower.7ex\hbox{E}\kern-.125emX}}
\usepackage{lipsum} % package to generate dummy text
\usepackage{stfloats} % package to force float positioning 

\begin{document}

\title{Weighted Sampled Split Learning (WSSL): Balancing Privacy, Robustness, and Fairness in Distributed Learning Environments}

\author{\IEEEauthorblockN{Manish Osti\IEEEauthorrefmark{1}, Aashray Thakuri \IEEEauthorrefmark {2}, Basheer Qolomany\IEEEauthorrefmark{3}, and Aos Mulahuwaish\IEEEauthorrefmark {1}}
\IEEEauthorblockA{\IEEEauthorrefmark{1} Department of Computer Science and Information Systems, Saginaw Valley State University, University Center, MI, USA}
\IEEEauthorblockA{\IEEEauthorrefmark{2} College of Engineering and Computer Science, University of Michigan- Dearborn, Dearborn, USA}
\IEEEauthorblockA{\IEEEauthorrefmark{3} Cyber Systems Department, University of Nebraska at Kearney, Kearney, USA}}

\maketitle

\begin{abstract}
This study presents Weighted Sampled Split Learning (WSSL), an innovative framework tailored to bolster privacy, robustness, and fairness in distributed machine learning systems. Unlike traditional approaches, WSSL disperses the learning process among multiple clients, thereby safeguarding data confidentiality. Central to WSSL's efficacy is its utilization of weighted sampling. This approach ensures equitable learning by tactically selecting influential clients based on their contributions. Our evaluation of WSSL spanned various client configurations and employed two distinct datasets: Human Gait Sensor and CIFAR-10. We observed three primary benefits: heightened model accuracy, enhanced robustness, and maintained fairness across diverse client compositions. Notably, our distributed frameworks consistently surpassed centralized counterparts, registering accuracy peaks of 82.63\% and 75.51\% for the Human Gait Sensor and CIFAR-10 datasets, respectively. These figures contrast with the top accuracies of 81.12\% and 58.60\% achieved by centralized systems. Collectively, our findings champion WSSL as a potent and scalable successor to conventional centralized learning, marking it as a pivotal stride forward in privacy-focused, resilient, and impartial distributed machine learning.
\end{abstract}

\begin{IEEEkeywords}
Weighted Sampled Split Learning (WSSL), Privacy-Preserving Machine Learning, Unbiased Learning, Robust Learning, and Distributed Learning.
\end{IEEEkeywords}

\section{Introduction} \label{sec:introduction}

In today's world, characterized by data ubiquity, the imperative for privacy-preserving machine learning amidst an explosion of data has made distributed learning essential \cite{shokri2015privacy}. Split learning stands out among distributed techniques, allowing deep neural networks to train on decentralized data sources without compromising data privacy \cite{gupta2018distributed, poirot2019split, vepakomma2018split}. By dividing models into front-end and back-end portions, split learning allows client devices to handle initial processing. In contrast, servers take on the bulk of computational duties \cite{langer2020distributed}. This approach, which emphasizes sending intermediate model representations over raw data, offers numerous advantages. It's especially relevant in sectors with high privacy demands, such as healthcare and finance \cite{bonawitz2019towards, konevcny2016federated, konevcny2016federated1}. Moreover, this transmission strategy reduces communication burdens, making it bandwidth-efficient and particularly valuable in networks with limited connectivity.

Our contributions in this space are:

\begin{enumerate}
\item \textbf{WSSL Algorithm}: We present the Weighted Sampled Split Learning (WSSL) algorithm, spotlighting the role of data distribution in distributed learning. This approach tackles the challenge of uneven data distribution in decentralized contexts and proposes a technique to enhance model training efficiency.

\item \textbf{Dynamic Client Importance-based Selection}: Instead of conventional static client choices, our model employs a dynamic system that varies client participation based on their contribution to the learning. This adaptability ensures consistent training and leads to better model results. This method, paired with global model averaging, fosters cooperative learning among clients, yielding improved training performance.

\item \textbf{Privacy, Robustness, and Fairness in Depth}: WSSL advances beyond typical distributed learning frameworks. By blending inputs from various client models, our technique strengthens model robustness, reduces the risk of overfitting, and promotes impressive generalization. Our unique client selection algorithm, which determines weights from validation performance, ensures that all clients have a balanced influence on learning, maintaining fairness. Empirical studies validate our method's superiority over centralized systems in robustness, fairness, and privacy.

\item \textbf{Practical Implications and Performance Metrics}: Our findings have tangible real-world applications, especially in resource-limited settings. For example, using the Human Gait Sensor dataset, WSSL achieved an impressive accuracy of 82.63\% in a 2-client setting over 20 communication rounds. This surpassed the 81.12\% peak accuracy of centralized methods. When tested with CIFAR-10, our distributed approach consistently outdid its centralized counterpart. A 10-client setup, for instance, attained an accuracy of 75.51\%, significantly outpacing the centralized peak of 58.60\%. These results emphasize the potential and adaptability of WSSL in a range of dataset conditions.

\end{enumerate}

With these advances, our work moves beyond traditional distributed learning limits, emphasizing data distribution's critical role. It paves the way for a refined WSSL methodology set to redefine model training efficiency.

%\begingroup
%\titlespacing*{\section}{0pt}{\parskip}{0pt}

\section {Related Work} \label{sec:relatedWorks}

This section identifies the research related to our work, spanning different areas. We also pinpoint the existing research gap and elucidate how our approach addresses and bridges this gap.

\subsection{Split Learning in Healthcare}

Both \cite{poirot2019split} and \cite{vepakomma2018split} advocate for split learning, emphasizing its applicability in collaborative training on limited healthcare data with a focus on data privacy. The challenges of this approach become evident when \cite{vepakomma2019reducing} introduces a method aimed at curtailing data leakage in distributed health data learning. Furthermore, \cite{abuadbba2020can} identifies potential privacy leakages within split learning. Interestingly, this is observed even when the split learning models achieve similar accuracy as their non-split counterparts in tasks such as heart abnormality detection using 1D CNNs. While mitigation techniques have been explored, they appear to be insufficient in isolation. Collectively, these studies highlight the evolution of split learning and the imminent need to explore data privacy assurance further.

\subsection{Integration of Federated and Split Learning}

The synthesis of federated learning (FL) and split learning (SL) has been explored by multiple studies. \cite{thapa2022splitfed} presents splitfed learning (SFL), which combines the strengths of both FL and SL. This approach promises enhanced model privacy, quicker training, and accuracy levels on par with traditional SL. Delving deeper into SL's intricacies, \cite{thapa2021advancements} details its advantages and potential pitfalls, notably resource efficiency and data leakage. Similarly, \cite{abedi2020fedsl} showcases an architecture that seamlessly blends FL and SL, offering enhanced privacy, improved accuracy, and efficient communication for sequentially partitioned data. Together, these studies champion the harmonious fusion of FL and SL, emphasizing their collective strengths in ensuring privacy, boosting efficiency, and maintaining accuracy.

\subsection{IoT and Edge-device Machine Learning}

In the Internet of Things (IoT) context, \cite{gao2020end} compares SplitNN and FL. The findings suggest that while SplitNN excels in handling imbalanced data, it falters with extreme non-IID data. In contrast, FL emerges as a more fitting solution for IoT environments, especially when deploying 1D CNN models. However, both methodologies encounter challenges, especially when dealing with intricate models on resource-constrained devices. Broadening the scope of split learning, \cite{koda2019one} introduces a multi-modal framework tailored for enhancing mmWave RF signal power predictions. Additionally, \cite{lim2020incentive} shifts the focus to edge learning in 5G networks, underscoring resource efficiency and inventive incentive mechanisms. These contributions collectively reinforce the potential of split learning in IoT settings, promising heightened accuracy, stringent data privacy, optimal resource utilization, and collaborative incentives.

\subsection{Security and Efficiency in Distributed Learning}

Security concerns associated with split learning are brought to the fore by \cite{pasquini2021unleashing}. This study not only underscores potential vulnerabilities but also illustrates inference attacks on client data orchestrated by malicious servers. A detailed comparison between the communication efficiencies of split learning and federated learning is presented by \cite{singh2019detailed}. Their findings suggest that split learning takes the lead in scenarios characterized by larger client bases or expansive model sizes. In contrast, federated learning showcases superior performance with abundant data samples and more constrained client or model sizes. Further, \cite{vepakomma2018no} provides insights into distributed deep learning models, emphasizing those that operate without direct client data access, and evaluates the associated merits, challenges, and trade-offs.

\subsection{Identifying the Gap}

Despite the comprehensive insights provided by the aforementioned studies, a significant research gap is evident regarding the distribution of data in distributed learning, particularly concerning non-IID datasets. Many strides have been made in split learning and federated learning methodologies. However, the nuances of data distribution across various clients and the inherent challenges presented by non-IID datasets have often been overlooked or underemphasized. This oversight is especially salient in scenarios where data is intrinsically diverse and distributed unevenly across participants. The current literature is noticeably bereft of specialized methodologies that adeptly address these challenges.

It's this very gap that the proposed Weighted Sampled Split Learning (WSSL) intends to fill, positioning itself as a pioneering approach to refine distributed learning practices. By accentuating dynamic client selection based on importance weights, WSSL adeptly tackles the multifaceted issue of non-uniform data distribution in decentralized settings. This innovative strategy ensures that all client contributions to the global model training are optimal, regardless of any disparities in individual data distributions. Through careful design and consideration, WSSL aims to elevate the overall performance and efficiency of distributed learning systems, providing a framework that not only fills the existing gap but also promises to redefine best practices in the realm of distributed machine learning.

%\endgroup

\section{Proposed Framework} \label{sec:proposedFramework}

This section unveils our innovative framework, Weighted Sampling Split Learning (WSSL). This framework adeptly integrates the concepts of weighted sampling with split learning, promising to be a game-changer in distributed machine learning by amplifying its efficiency and performance. Central to WSSL is a re-envisioned model training process that disseminates tasks between a myriad of clients and a pivotal server. This distribution takes into account the significance of each client, astutely calibrating their respective contributions. The detailed mechanics and procedures of our proposed structure are delineated in the subsequent sections.

\subsection{System Model and Problem Objective}
We consider a distributed learning system consisting of a central server and multiple edge clients. In this setup, we aim to achieve a collaborative model training. A significant constraint is that the server does not have direct access to the raw data of the clients, which is essential to ensure data privacy and meet certain regulatory requirements.

Given:
\begin{itemize}
    \item \(S\) - Central server
    \item \(C_i, i=1,...,N\) - N clients 
    \item \(D_i\) - Local dataset for client \(C_i\)
    \item \(f_i()\) - Local model for client \(C_i\)
    \item \(L_i\) - Loss for client \(C_i\)
\end{itemize}

We aim to train individual client models \(f_i()\), such that their associated losses are minimized, and their outputs are consistent with a global model \(f_{\text{global}}()\) residing at the server. Mathematically, our objective can be formulated as follows:

\[ \min_{f_1,...,f_N} \sum_{i=1}^N L_i(f_i(D_i), Y) \]
Subject to:
\[ f_i(D_i) = f_{\text{global}}(D) \]

Where \(f_{\text{global}}()\) represents the global model at the server.

\subsection{Importance-based Client Selection}

In the Weighted Sampling Split Learning (WSSL) framework, client selection for each training round is paramount. Traditional federated learning often neglects variations in data distributions, data quality, and past performance across different clients, treating them uniformly. In contrast, our method employs an importance-based weighting scheme to differentiate clients, ensuring the most suitable clients are selected for model refinement. To operationalize this philosophy into a systematic approach, we introduce the Weighted Sampled Split Learning (WSSL) Client Selection Algorithm.

\begin{enumerate}
\item \textbf{Weighted Sampled Split Learning (WSSL) Client Selection Algorithm (Algorithm~\ref{alg:WSSLClient}):} This algorithm outlines the specifics of our importance-based selection. The central server employs it to handpick a set of clients for each epoch, focusing on those that have the greatest potential to refine the global model. The steps of this algorithm are as follows:

    \begin{enumerate}
        \item \textbf{Initialization:} Begin by setting the total number of clients, denoted as $\alpha$, and iterate through the epochs.
        
        \item \textbf{Importance Weight Calculation:} For every epoch, the server calculates the importance weight ($\beta_i$) of each client using the function \textit{compute importance}. This function incorporates factors such as data quality, alignment with the global model, and past performance to determine the potential impact of a client.
        
        \item \textbf{Normalization and Client Selection:} Normalize the importance weights to produce $\gamma_i$, ensuring their sum equals 1. These normalized weights quantify the likelihood of each client's valuable contribution in the current epoch. The number of clients to engage, represented by $\alpha'$, is deduced from the normalized weights' weighted average, with at least one client selected to maintain diversity.
        
        \item \textbf{Weighted Sampling:} Employing principles from probabilistic sampling, clients with elevated normalized weights are more likely to be selected. The list, $\delta$, encapsulates the chosen clients who will partake in the next training iteration.
    \end{enumerate}
    
    \item \textbf{Integrating Algorithm~\ref{alg:WSSLClient} into WSSL,} this importance-driven client selection approach not only ensures fairness by recognizing unique client attributes but also augments the proficiency of the global model. By attributing distinct importance weights that resonate with each client's data quality, expertise, and past performance, WSSL fosters a balanced, data-centric learning strategy, amplifying the merits of distributed machine learning.
\end{enumerate}

\begin{algorithm}[ht]
\scriptsize
\caption{Weighted Sampled Split Learning (WSSL) Client Selection}\label{alg:WSSLClient}
\textbf{Notations:} $\alpha$ = Total number of clients, $\beta$ = Importance weights, $\gamma$ = Normalized weights, $\delta$ = Selected clients, $\epsilon$ = Server model, $\zeta$ = Validation dataset.
\begin{algorithmic}[1]
\State Initialize: $\alpha \leftarrow$ Total number of clients.
\For{each $i$ in number of epochs}
\If{$i == 0$}
\State $\delta \leftarrow [0, 1, ..., \alpha-1]$
\Else
\State Compute importance weights of each client: $\beta_{i}$ $\leftarrow$ compute importance$(client_{i}, \epsilon, \zeta)$
\State Compute sum of weights: $\beta_{sum}$ $\leftarrow$ $\sum_{i=1}^{\alpha}\beta_{i}$
\State Normalize weights: $\gamma_{i}$ $\leftarrow$ $\frac{\beta_{i}}{\beta_{sum}}$
\State Determine number of clients to select: $\alpha^{}$ $\leftarrow$ $max(\alpha \times average(\gamma), 1)$
\State Handpick clients: $\delta$ $\leftarrow$ weighted sampling $([0, 1, ..., \alpha-1], \gamma_{i}, \alpha^{})$
\EndIf
\EndFor
\end{algorithmic}
\end{algorithm}

\subsection{Training and Model Updates}

The Weighted Sampling Split Learning (WSSL) architecture stands as a testament to the power of collaboration. Seamlessly weaving the central server with individual clients, it showcases a harmonized dance of training processes and model updates, detailed in Algorithm~\ref{alg:WSSL}. More than just an exchange, this methodology amplifies knowledge enhancement throughout the system. Let's delve deeper:

\begin{enumerate}
\item \textbf{Algorithm~\ref{alg:WSSL} - Weighted Sampled Split Learning (WSSL):} Shedding light on the WSSL framework's training mechanics, this algorithm underscores the synchronized endeavors of the server and client entities, all striving towards refining the global model. The steps are as follows:

\begin{enumerate}
    \item \textbf{Importance Computation and Sampling:} Each epoch kickstarts with the server determining the importance weights ($w[i]$) for every client, guided by Algorithm~\ref{alg:WSSLClient}. Based on these weights, clients are chosen for the current training round, ensuring that influential ones play a central role in global optimization.
    
    \item \textbf{Client-Side Training:} Selected clients embark on local training using their datasets. As their model ($C[i]$) parameters ($\theta[i]$) adapt through gradient descent, their intermediate representations ($a[i]$) are sent to the server for global integration.
    
    \item \textbf{Server-Side Training:} Armed with the intermediate representations ($a[i]$), the server evaluates the loss ($L[i]$) for each client and processes gradients via backpropagation. These gradients are subsequently dispatched to the corresponding clients.
    
    \item \textbf{Client-Side Update:} Clients, upon receiving gradients from the server, refine their model parameters, aligning more closely with global objectives.
    
    \item \textbf{Global Model Update:} By amalgamating all client contributions, the server computes a weighted average of their parameters ($\theta_{\text{global}}$), encapsulating the shared wisdom.
\end{enumerate}

\item \textbf{Stability of Importance Weights:} Beyond the algorithm's iterative updates, it's essential to underscore the consistent nature of importance weights. Contrasting dynamically updated model parameters, these weights offer stability across training rounds, ensuring undeterred influence over the learning curve.

\end{enumerate}

For a visually enriched grasp of this distributed machine learning paradigm, readers are directed to Figure 1. This methodology, which orchestrates a dual backpropagation between client and server models, is further exemplified using two case studies: the human gait sensor and the CIFAR-10 dataset, dissected further in Section~\ref{sec:datasets}.

\subsection{Addressing Trustworthiness}
To bolster trustworthiness within our framework, we integrate the following measures:
\begin{enumerate}
    \item \textbf{Data Integrity}: Prior to training, a hashing mechanism is implemented on the client's data. This not only ensures the preservation of data privacy but also allows the server to verify data integrity without accessing the raw data.
    
    \item \textbf{Model Validity}: After training, the server can optionally deploy techniques like model watermarking. This ensures that the model updates retrieved from clients are authentic and remain unaltered.
\end{enumerate}

\subsection{Comparison with Existing Methods}
What makes our approach superior is twofold:
\begin{enumerate}
    \item \textbf{Fairness in Client Participation}: Unlike traditional methods that randomly select clients or prioritize clients with larger datasets, our approach guarantees fairness by giving every client a chance to be selected based on their importance.
    \item \textbf{Efficiency in Communication}: By focusing on important clients and not mandating participation from all, we reduce the communication overhead, making training faster.
\end{enumerate}

\begin{algorithm}
\scriptsize
\caption{Weighted Sampled Split Learning (WSSL)}\label{alg:WSSL}
\textbf{Notations:} 
\begin{enumerate}
    \item \(C[i]\) - the $i-th$ client model,
    \item \(S\) - server model,
    \item \(D[i]\) - local dataset for client \(C[i]\),
    \item \(\omega\) - parameters of server model \(S\),
    \item \(\theta[i]\) - parameters of the \(i-th\) client model \(C[i]\),
    \item \(a[i]\) - intermediate representation from client model \(C[i]\),
    \item \(L[i]\) - loss for the \(i-th\) client,
    \item \(w[i]\) - importance weights for the \(i-th\) client,
    \item \(\theta_{\text{global}}\) - weighted average of client models' parameters.
\end{enumerate}
\textbf{Input:} Local datasets \(D[i]\) for each client \(C[i]\), server model \(S\) with parameters \(\omega\), client models \(C[i]\) with parameters \(\theta[i]\).
\textbf{Output:} Updated server model \(S\) and client models \(C[i]\).

\begin{algorithmic}[1]
\State \textbf{Procedure:}

\State \underline{\textbf{1. Compute Importance and Sampling:}}
\For{each client \(C[i]\)}
    \State Compute \(w[i]\) to represent \(C[i]\)'s importance.
    \State Normalize weights: \(w[i] \leftarrow \frac{w[i]}{\sum(w)}\).
    \State Select \(C[i]\) for current training round based on its normalized weight.
\EndFor

\State \underline{\textbf{2. Client-Side Training:}}
\For{selected client \(C[i]\) in the current round}
    \State Start with initial parameters \(\theta[i]\).
    \For{each data batch \((x, y)\) from \(D[i]\)}
        \State Obtain \(a[i] = C[i](x)\).
        \State Detach \(a[i]\) from computation graph and forward to server.
    \EndFor
\EndFor

\State \underline{\textbf{3. Server-Side Training:}}
\State Receive \(a[i]\) from each active client \(C[i]\).
\For{client \(C[i]\) sending \(a[i]\)}
    \State Calculate loss \(L[i] = L(S(a[i]), y)\).
    \State Obtain gradients \(\frac{\partial L[i]}{\partial a[i]}\) via backpropagation on \(S\).
    \State Return gradients to respective \(C[i]\).
\EndFor

\State \underline{\textbf{4. Client-Side Model Update:}}
\For{client \(C[i]\) receiving server gradients}
    \State Update \(C[i]\) using received \(\frac{\partial L[i]}{\partial a[i]}\).
    \State Refine \(\theta[i]\) using gradient descent.
\EndFor

\State \underline{\textbf{5. Global Model Update:}}
\State Compute:
\[ \theta_{\text{global}} = \sum_i w[i] \times \theta[i] \]
\State \textbf{Return:} Updated server model \(S\) and client models \(C[i]\).
\end{algorithmic}
\end{algorithm}

\begin{figure}[ht] 
    \centering
    \includegraphics[width=0.8\linewidth]{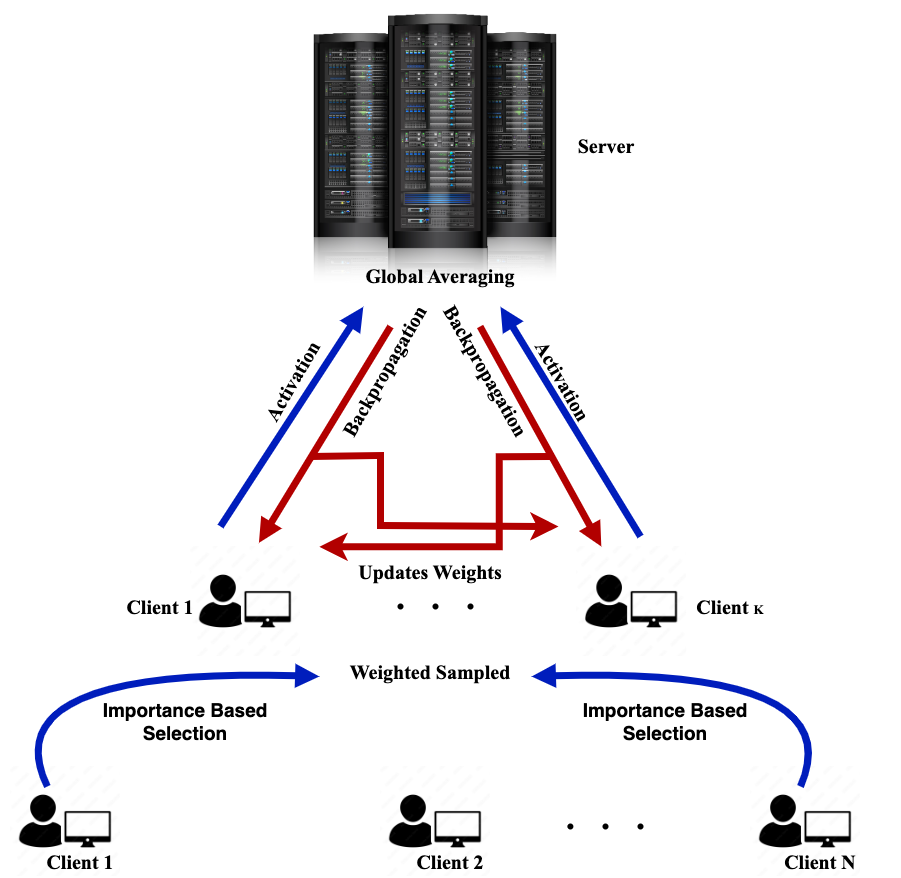}
    \caption{Overview of Weighted Sampling Split Learning System}
    \label{fig:fig1}
\end{figure}

\section {Datasets} \label{sec:datasets}

The subsequent subsections provide a description of the datasets utilized for experimental evaluations and the preprocessing steps applied to the data. The research employs two distinct datasets, namely the Human Gait Sensor dataset \cite {luo2020database} and the CIFAR-10 dataset \cite {krizhevsky2009learning}, to showcase the adaptability and robustness of Weighted Sampling Split Learning (WSSL). The Human Gait sensor dataset demonstrates WSSL's potential in processing real-world sensor data, particularly in domains such as health monitoring and IoT, while the CIFAR-10 dataset evaluates its performance in image data classification, highlighting its applicability to vision-based tasks. Through the use of diverse datasets, this research offers a comprehensive evaluation of WSSL's performance in distributed machine learning scenarios, underscoring its broad applicability and effectiveness.

\subsection{Human Gait Sensor Case Study}
We utilized a comprehensive dataset comprising 2,803,999 observations, with each observation containing 28 attributes derived from various sensor readings. The dataset used in this research study comprises 30 participants, with an equal distribution of 15 males and 15 females, ensuring gender balance and unbiased representation. In the preprocessing stage, the dataset was segregated into features and labels (Gender), enabling supervised learning. The features matrix encompassed all columns except the target variable, which served as the label. Following this, an 80-20 split strategy was employed, allocating 80\% of the data for model training and reserving the remaining 20\% for model validation and testing. This division ensured that the model had a significant amount of data to learn from while providing an independent set for evaluating its generalization. To mitigate any disparities in the magnitude of values across columns, standard scaling was applied to normalize the data. This procedure transformed the features to have a mean of 0 and a standard deviation of 1, ensuring a consistent scale and preventing dominant influences of variables with larger values. Finally, the data was converted into PyTorch tensors, a crucial step considering the reliance on PyTorch, a widely used open-source machine learning library, for subsequent model building and training stages.

\subsection{CIFAR-10 Case Study}
The dataset used in this study is the CIFAR-10 dataset, which consists of 60,000 color images of 32x32 pixels, categorized into 10 distinct classes such as airplane, automobile, bird, cat, deer, dog, frog, horse, ship, and truck. The dataset is split into a training set of 50,000 images and a test set of 10,000 images. The data undergoes preprocessing, including conversion to tensors and normalization, to facilitate learning and convergence. To partition the dataset for each client in our weighted sampled split learning approach, we employ a stratified sampling technique, ensuring that each client's subset maintains the same class distribution as the original dataset. Each client is then provided with a data loader, allowing for batch processing and data shuffling to expose the model to diverse samples and minimize overfitting.

\section{Experimental Setup and Evaluation} \label{sec:experimentsResulsts}

\subsection{System Specifications}
Our experimental environment was provisioned on a machine boasting 32 GB of RAM paired with a 6 GB Nvidia 2060 RTX GPU. For crafting and training the neural network models, we employed PyTorch, a renowned deep learning framework prized for its comprehensive ecosystem and intuitive design capabilities.

\subsection{Centralized Training}
Contrastingly different from the split learning approach, centralized training amasses all data and computational muscle on one unified server. Here, the entirety of the neural network model gets trained to utilize centrally stored data. Though it provides the luxury of accessing the complete data simultaneously, this method could grapple with scalability, data privacy, and bandwidth consumption.

For clarity, the centralized systems in our experiments retain the unabridged architectures of the models, eliminating any division between client and server roles. Their specifications can be further consulted in Table \ref{tab:table1}.

\subsection{Datasets and Model Architectures}

\subsubsection{Human Gait Sensor Dataset}
The dataset incorporated a dual-segment feedforward neural network design:

\begin{itemize}
    \item \textbf{Client-Side Model:} Positioned on the edge device, this segment is a simple fully connected neural network spanning two layers. Post accepting the input data, it carries out a linear transformation (leveraging layer-specific weights and biases) and applies the ReLU activation function. Its second layer's output acts as the intermediate representation, forwarded to the server model.
    
    \item \textbf{Server-Side Model:} This counterpart is designed for the server, which processes the intermediate data sent by the client model. After undergoing a linear transformation and ReLU activation across its first two layers, the final layer employs a sigmoid activation function—ideal for binary classification challenges—to restrict the output between 0 and 1.
\end{itemize}

\subsubsection{CIFAR-10 Dataset}
For this dataset, a variant of the famed ResNet-18 convolutional neural network architecture was adapted, split strategically for both client and server sides to classify 10 distinct image classes:

\begin{itemize}
    \item \textbf{Client-Side Model:} Occupying the early layers of the ResNet-18 architecture up to a defined cut-off, this segment primarily transmutes raw input into intermediate feature sets, dispatched to the server model.
    
    \item \textbf{Server-Side Model:} Situated on the server, this portion picks up from the client-defined cut-off and stretches till the ResNet-18's final layer. It processes the intermediate data, culminating in the final output. Its terminal layer is a fully connected layer, followed by a Softmax function, fitting for rendering class probabilities in multi-class challenges.
\end{itemize}

The cut-off point becomes a pivotal element, balancing the client-side computational demands with the data transmission volume to the server. Adjusting it judiciously is crucial, factoring in the client device's computational prowess and network bandwidth.

\subsection{Model Evaluation Strategy}
For rigorous model evaluations, the StratifiedShuffleSplit function from the scikit-learn library was our tool of choice. By ensuring a proportional representation of each class in every fold, this cross-validation technique is indispensable for skewed datasets. This method promotes a balanced and robust data split, augmenting the veracity of our evaluation metrics.

Lastly, Table \ref{tab:table1} enumerates the architectural nuances of the models in our experiments, juxtaposing the centralized frameworks with the weighted sampled split learning configurations.

\subsection{Human Gait Sensor Evaluation Results}
\begin{figure*}[t] 
  \centering
 
  \begin{subfigure}[t]{0.45\textwidth}
    \includegraphics[width=\textwidth]{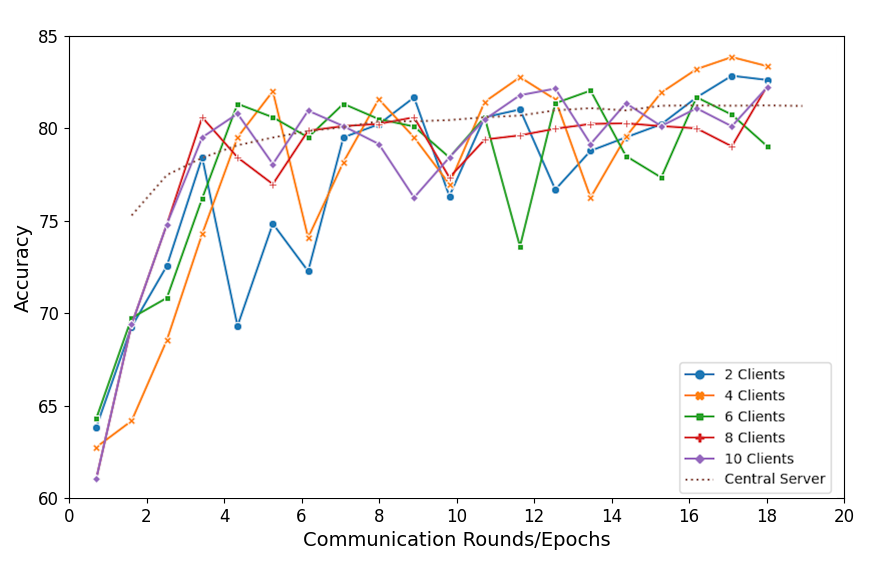}
    \caption{Accuracy vs. Communication Rounds for Human Gait Sensor Data }
    \label{subfig:a}
  \end{subfigure}
  \hfill
  \begin{subfigure}[t]{0.45\textwidth}
    \includegraphics[width=\textwidth]{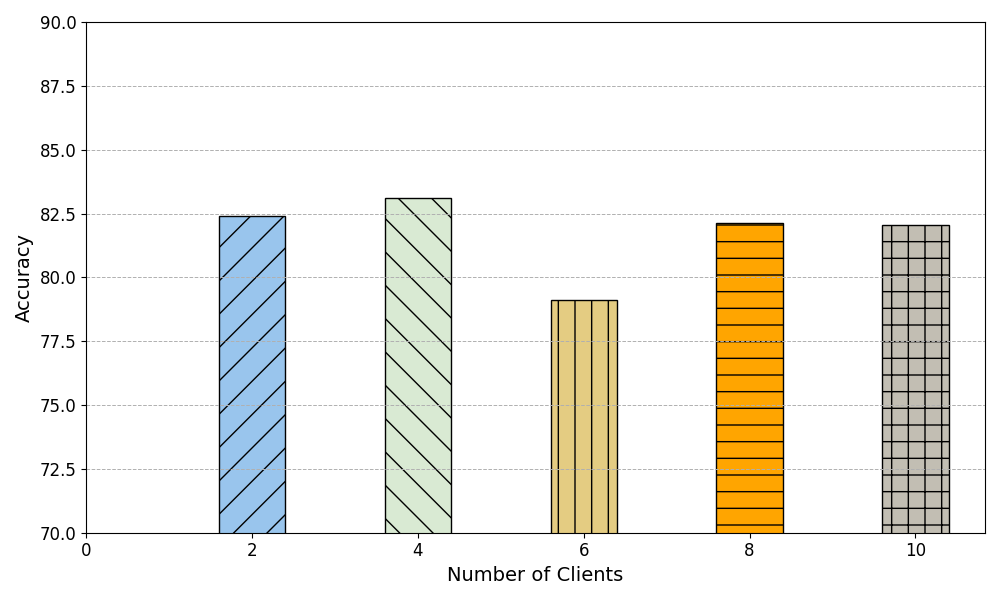}
    \caption{Accuracy vs. Number of Clients for Human Gait Sensor Data}
    \label{subfig:b}
  \end{subfigure}
  
  \vspace{0cm} % Add some vertical space between the rows
  
  \begin{subfigure}[t]{0.45\textwidth}
    \includegraphics[width=\textwidth]{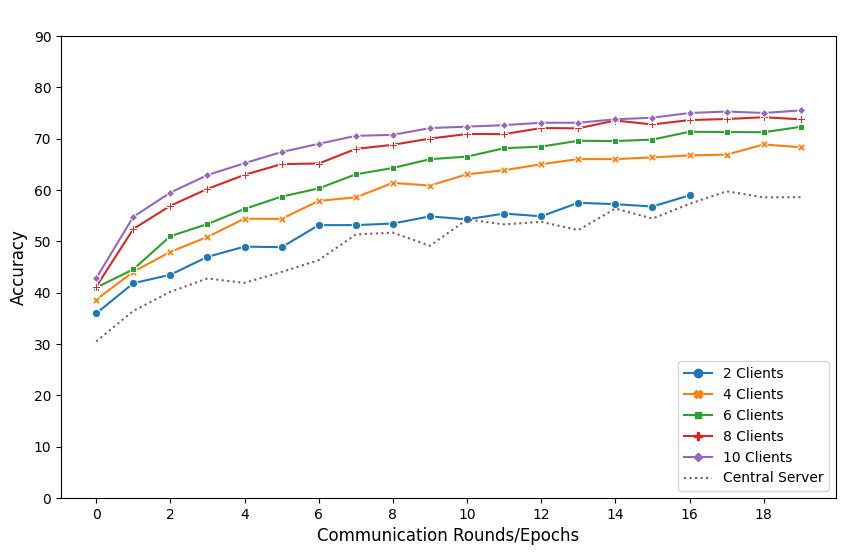}
    \caption{Accuracy vs. Communication Rounds for CIFAR-10 Data }
    \label{subfig:c}
  \end{subfigure}
  \hfill
  \begin{subfigure}[t]{0.45\textwidth}
    \includegraphics[width=\textwidth]{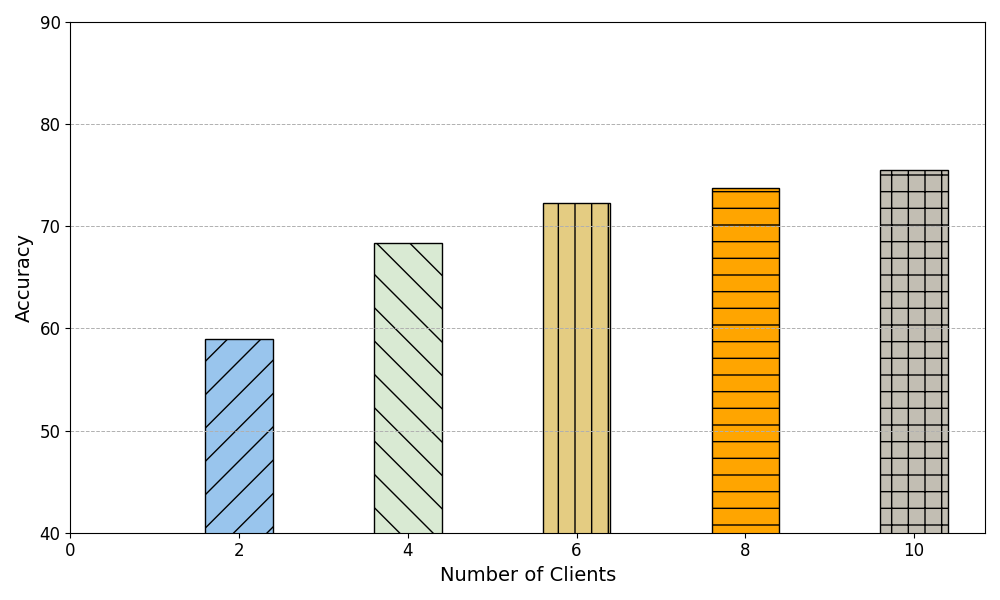}
    \caption{Accuracy vs. Number of Clients for CIFAR-10 Data}
    \label{subfig:d}
  \end{subfigure}
  
  \caption{Accuracy vs. Communication Rounds and Number of Clients for the Two Datasets}
  \label{fig:fig2}
\end{figure*}

We gauged the efficiency of the weighted sampled split learning approach on the Human Gait Sensor dataset by varying the number of clients from 2 to 10 across 20 communication rounds or epochs. With each configuration, a significant uptrend in performance was evident. As displayed in Figure \ref{subfig:a} and \ref{subfig:b}, the system initialized with 2 clients began at an accuracy of 65.12\%, surging to 82.63\% by the 20th epoch. Comparable upward trends materialized for setups with 4, 6, 8, and 10 clients, culminating in accuracies of 83.11\%, 79.12\%, 82.12\%, and 82.06\%, respectively. Notably, the 4, 8, and 10 client systems surpassed the centralized learning's peak accuracy of 81.12\%. These observations underline the potency and scalability of our distributed approach, which not only thrives with an increased client base but also provides a significant edge over conventional centralized learning.

\begin{table*}[!t]
    \centering
    \begin{tabular}{|c|c|c|c|c|c|}
        \hline
        \textbf{Architecture} & \textbf{Dataset} & \textbf{\#Parameters} & \textbf{Layers} & \textbf{Kernel Size} & \textbf{Batch Size} \\
        \hline
        Feedforward Neural Network & Human Gait Sensor & 32,000.00 & 5 & - & 128 \\
        \hline
        ResNet18 & CIFAR-10 & 11.7 million & 18 &(7 × 7), (3 × 3) & 128 \\
        \hline
    \end{tabular}
    \caption{Models' Architecture}
    \label{tab:table1}
\end{table*}

%\lipsum[3-10] % Generate some more dummy text

\subsection{CIFAR-10 Evaluation Results}
The CIFAR-10 dataset served as another testbed for our weighted sampled split learning system. As portrayed in Figure \ref{subfig:c} and \ref{subfig:d}, the model, when started with 2 clients, had an inception accuracy of 35.96\%, which rose to 58.96\% by the concluding epoch. Systems incorporating 4, 6, 8, and 10 clients showcased similar growth patterns, ending with accuracies of 68.34\%, 72.30\%, 73.77\%, and 75.51\%, respectively. For perspective, the traditional centralized approach commenced at 30.50\% and plateaued at 58.60\% by the 20th epoch. Crucially, every distributed setup with varying client counts eclipsed the centralized model in performance. These findings bolster the case for weighted sampled split learning, especially when considering its heightened adaptability as client numbers grow. This trend remained consistent even with the intricacies of the CIFAR-10 dataset, where our method continually overshadowed the standard centralized learning paradigm.

\section{Findings and Lessons Learned}

Through detailed analysis and exploration of Weighted Sampled Split Learning, we distilled the following vital insights:

\begin{enumerate}

\item \textbf{Robustness and Fairness}: Our study underlines the enhanced robustness and fairness offered by weighted sampled split learning compared to conventional centralized methods. The strategic weighting prevents undue dominance by specific datasets or clients, fostering a balanced environment where every client's contributions are valued and recognized.

\item \textbf{Privacy and Accuracy}: Weighted sampled split learning adeptly navigates the delicate balance between data privacy and model performance. By confining data to its original location and only exchanging model updates, privacy is upheld. The amalgamated efforts of the clients, shaped by their importance weights, fortify the model's precision without compromising its reliability.

\end{enumerate}

These insights validate weighted sampled split learning as a powerful approach in distributed machine learning, adeptly turning challenges into assets.

\section{Conclusions and Future Directions} \label{sec:conclusions}

In this work, we've spotlighted Weighted Sampled Split Learning (WSSL) as a groundbreaking approach to addressing the intricacies of privacy, robustness, and fairness in decentralized machine learning environments.

From our research, we highlight:

\begin{enumerate}

\item \textbf{Enhanced Model Performance}: WSSL, through importance-based client selection and harmonized learning, consistently outperforms traditional centralized models.

\item \textbf{Upholding Privacy and Fairness}: WSSL inherently protects individual data and champions equitable client participation, marking a significant leap in distributed learning.

\item \textbf{Sturdy Decentralized Learning}: The strength of WSSL lies in its combined emphasis on cooperative learning and prioritized client involvement.

\end{enumerate}

Looking forward, intriguing areas of exploration are:

\begin{enumerate}

\item \textbf{Client Dynamics and Weighting Impact}: Delving deeper into the interplay between the number of clients and their respective weightings.

\item \textbf{Addressing Imbalances and Scalability}: Adapting WSSL to scenarios with imbalanced data and scaling to expansive environments.

\item \textbf{Refining Split Points}: Fine-tuning the junctures in neural network structures to adeptly navigate between localized and centralized processing.

\end{enumerate}

In essence, WSSL offers a pioneering pathway in distributed machine learning. With its demonstrated advantages and the potential it opens up for further innovation, it stands poised to redefine decentralized learning, emphasizing privacy, robustness, and fairness.

\bibliographystyle{IEEEtran}
\bibliography{sample}

\begin{thebibliography}{10}
\providecommand{\url}[1]{#1}
\csname url@samestyle\endcsname
\providecommand{\newblock}{\relax}
\providecommand{\bibinfo}[2]{#2}
\providecommand{\BIBentrySTDinterwordspacing}{\spaceskip=0pt\relax}
\providecommand{\BIBentryALTinterwordstretchfactor}{4}
\providecommand{\BIBentryALTinterwordspacing}{\spaceskip=\fontdimen2\font plus
\BIBentryALTinterwordstretchfactor\fontdimen3\font minus
  \fontdimen4\font\relax}
\providecommand{\BIBforeignlanguage}[2]{{%
\expandafter\ifx\csname l@#1\endcsname\relax
\typeout{** WARNING: IEEEtran.bst: No hyphenation pattern has been}%
\typeout{** loaded for the language `#1'. Using the pattern for}%
\typeout{** the default language instead.}%
\else
\language=\csname l@#1\endcsname
\fi
#2}}
\providecommand{\BIBdecl}{\relax}
\BIBdecl

\bibitem{shokri2015privacy}
R.~Shokri and V.~Shmatikov, ``Privacy-preserving deep learning,'' in
  \emph{Proceedings of the 22nd ACM SIGSAC conference on computer and
  communications security}, 2015, pp. 1310--1321.

\bibitem{gupta2018distributed}
O.~Gupta and R.~Raskar, ``Distributed learning of deep neural network over
  multiple agents,'' \emph{Journal of Network and Computer Applications}, vol.
  116, pp. 1--8, 2018.

\bibitem{poirot2019split}
M.~G. Poirot, P.~Vepakomma, K.~Chang, J.~Kalpathy-Cramer, R.~Gupta, and
  R.~Raskar, ``Split learning for collaborative deep learning in healthcare,''
  \emph{arXiv preprint arXiv:1912.12115}, 2019.

\bibitem{vepakomma2018split}
P.~Vepakomma, O.~Gupta, T.~Swedish, and R.~Raskar, ``Split learning for health:
  Distributed deep learning without sharing raw patient data,'' \emph{arXiv
  preprint arXiv:1812.00564}, 2018.

\bibitem{langer2020distributed}
M.~Langer, Z.~He, W.~Rahayu, and Y.~Xue, ``Distributed training of deep
  learning models: A taxonomic perspective,'' \emph{IEEE Transactions on
  Parallel and Distributed Systems}, vol.~31, no.~12, pp. 2802--2818, 2020.

\bibitem{bonawitz2019towards}
K.~Bonawitz, H.~Eichner, W.~Grieskamp, D.~Huba, A.~Ingerman, V.~Ivanov,
  C.~Kiddon, J.~Kone{\v{c}}n{\`y}, S.~Mazzocchi, B.~McMahan \emph{et~al.},
  ``Towards federated learning at scale: System design,'' \emph{Proceedings of
  machine learning and systems}, vol.~1, pp. 374--388, 2019.

\bibitem{konevcny2016federated}
J.~Kone{\v{c}}n{\`y}, H.~B. McMahan, D.~Ramage, and P.~Richt{\'a}rik,
  ``Federated optimization: Distributed machine learning for on-device
  intelligence,'' \emph{arXiv preprint arXiv:1610.02527}, 2016.

\bibitem{konevcny2016federated1}
J.~Kone{\v{c}}n{\`y}, H.~B. McMahan, F.~X. Yu, P.~Richt{\'a}rik, A.~T. Suresh,
  and D.~Bacon, ``Federated learning: Strategies for improving communication
  efficiency,'' \emph{arXiv preprint arXiv:1610.05492}, 2016.

\bibitem{vepakomma2019reducing}
P.~Vepakomma, O.~Gupta, A.~Dubey, and R.~Raskar, ``Reducing leakage in
  distributed deep learning for sensitive health data,'' \emph{arXiv preprint
  arXiv:1812.00564}, vol.~2, 2019.

\bibitem{abuadbba2020can}
S.~Abuadbba, K.~Kim, M.~Kim, C.~Thapa, S.~A. Camtepe, Y.~Gao, H.~Kim, and
  S.~Nepal, ``Can we use split learning on 1d cnn models for privacy preserving
  training?'' in \emph{Proceedings of the 15th ACM Asia Conference on Computer
  and Communications Security}, 2020, pp. 305--318.

\bibitem{thapa2022splitfed}
C.~Thapa, P.~C.~M. Arachchige, S.~Camtepe, and L.~Sun, ``Splitfed: When
  federated learning meets split learning,'' in \emph{Proceedings of the AAAI
  Conference on Artificial Intelligence}, vol.~36, no.~8, 2022, pp. 8485--8493.

\bibitem{thapa2021advancements}
C.~Thapa, M.~A.~P. Chamikara, and S.~A. Camtepe, ``Advancements of federated
  learning towards privacy preservation: from federated learning to split
  learning,'' \emph{Federated Learning Systems: Towards Next-Generation AI},
  pp. 79--109, 2021.

\bibitem{abedi2020fedsl}
A.~Abedi and S.~S. Khan, ``Fedsl: Federated split learning on distributed
  sequential data in recurrent neural networks,'' \emph{arXiv preprint
  arXiv:2011.03180}, 2020.

\bibitem{gao2020end}
Y.~Gao, M.~Kim, S.~Abuadbba, Y.~Kim, C.~Thapa, K.~Kim, S.~A. Camtepe, H.~Kim,
  and S.~Nepal, ``End-to-end evaluation of federated learning and split
  learning for internet of things,'' \emph{arXiv preprint arXiv:2003.13376},
  2020.

\bibitem{koda2019one}
Y.~Koda, J.~Park, M.~Bennis, K.~Yamamoto, T.~Nishio, and M.~Morikura, ``One
  pixel image and rf signal based split learning for mmwave received power
  prediction,'' in \emph{Proceedings of the 15th International Conference on
  emerging Networking EXperiments and Technologies}, 2019, pp. 54--56.

\bibitem{lim2020incentive}
W.~Y.~B. Lim, J.~S. Ng, Z.~Xiong, D.~Niyato, C.~Leung, C.~Miao, and Q.~Yang,
  ``Incentive mechanism design for resource sharing in collaborative edge
  learning,'' \emph{arXiv preprint arXiv:2006.00511}, 2020.

\bibitem{pasquini2021unleashing}
D.~Pasquini, G.~Ateniese, and M.~Bernaschi, ``Unleashing the tiger: Inference
  attacks on split learning,'' in \emph{Proceedings of the 2021 ACM SIGSAC
  Conference on Computer and Communications Security}, 2021, pp. 2113--2129.

\bibitem{singh2019detailed}
A.~Singh, P.~Vepakomma, O.~Gupta, and R.~Raskar, ``Detailed comparison of
  communication efficiency of split learning and federated learning,''
  \emph{arXiv preprint arXiv:1909.09145}, 2019.

\bibitem{vepakomma2018no}
P.~Vepakomma, T.~Swedish, R.~Raskar, O.~Gupta, and A.~Dubey, ``No peek: A
  survey of private distributed deep learning,'' \emph{arXiv preprint
  arXiv:1812.03288}, 2018.

\bibitem{luo2020database}
Y.~Luo, S.~M. Coppola, P.~C. Dixon, S.~Li, J.~T. Dennerlein, and B.~Hu, ``A
  database of human gait performance on irregular and uneven surfaces collected
  by wearable sensors,'' \emph{Scientific data}, vol.~7, no.~1, p. 219, 2020.

\bibitem{krizhevsky2009learning}
A.~Krizhevsky, G.~Hinton \emph{et~al.}, ``Learning multiple layers of features
  from tiny images,'' 2009.

\end{thebibliography}

\end{document}